# INCONSISTENT AFFECTIVE REACTION

*Sentiment of Perception and Opinion in Urban Environments*

JINGFEI HUANG[1] and HAN TU[2]
[1]*Harvard University,* [2]*Massachusetts Institute of Technology.*
[1]*jingfeihuang@mde.harvard.edu, 0009-0002-0213-4160*
[2]*hantu@mit.edu, 0000-0003-2525-8414*

**Abstract.** The ascension of social media platforms has transformed our understanding of urban environments, giving rise to nuanced variations in sentiment reaction embedded within human perception and opinion, and challenging existing multidimensional sentiment analysis approaches in urban studies. This study presents novel methodologies for identifying and elucidating sentiment inconsistency, constructing a dataset encompassing 140,750 Baidu and Tencent Street view images to measure perceptions, and 984,024 Weibo social media text posts to measure opinions. A reaction index is developed, integrating object detection and natural language processing techniques to classify sentiment in Beijing Second Ring for 2016 and 2022. Classified sentiment reaction is analysed and visualized using regression analysis, image segmentation, and word frequency based on land-use distribution to discern underlying factors. The perception affective reaction trend map reveals a shift toward more evenly distributed positive sentiment, while the opinion affective reaction trend map shows more extreme changes. Our mismatch map indicates significant disparities between the sentiments of human perception and opinion of urban areas over the years. Changes in sentiment reactions have significant relationships with elements such as dense buildings and pedestrian presence. Our inconsistent maps present perception and opinion sentiments before and after the pandemic and offer potential explanations and directions for environmental management, in formulating strategies for urban renewal.

## 1. Introduction

The ascendance of social media platforms heralds a transformation in the conceptualization of the physical milieu (Beyes et al., 2019; McLuhan, 1964; Oungrinis, 2006; Shook & Turner, 2016; Tao et al., 2019; Winner, 2023), engendering subtle variances in affections of human perception and opinion toward the urban environment (Becken et al., 2017; Beyes et al., 2019; Tao et al., 2019). Emerging from this shift is an epiphenomenon termed affective reaction inconsistency, manifested within physical urban environments characterized by ubiquitous social media use. This mismatch occurs when the affective reaction stimulated by the perception of physical environments differs from opinions expressed on social media. Specifically, people perceive environments through their preferred, albeit biased, digital lenses for posts on social media (van der Meer & Hameleers, 2022). This inconsistency challenges previous research (Becken et al., 2017; van der Meer & Hameleers, 2022; Yazıcıoğlu, 2017) on multidimensional affective reaction analysis of urban environments using social media data. Our study introduces methodologies to measure when and why affective reactions mismatch, as supplementary information for urban research, urban design, and urban management (Becken et al., 2017; Biljecki & Ito, 2021; Dai & Hua, 2019; Naik et al., 2014; Tao et al., 2019) using street view data and social media data.

Studies have focused on affective reactions expressed visually or textually within urban environments. Researchers have harnessed street view image analysis to elucidate the multifaceted factors influencing human reactions in physical settings. These factors encompass diverse dimensions, including green ratio (Pei et al., 2019), sky openness ratio, safety (Naik et al., 2014), perception scales of safety, liveliness, aesthetics, wealth, depression, and boringness (F. Zhang et al., 2018), as well as mixed factors (Dai & Hua, 2019; Ye et al., 2021). Other approaches seek to extract public responses using social media data from several perspectives, including air quality (Tao et al., 2019), tourist satisfaction (Alaei et al., 2017), and geolocations of social media posts (Becken et al., 2017). However, the stimulation people perceived from the physical environments and what they expressed on social media may differ, resulting in biased analysis and evaluations of urban environments. The absence of integrated analyses hinders a comparison of affective reactions between human perceptions and opinions. Therefore, our conceptual goal is to gain a more comprehensive view of how people feel about the urban space by bridging perception and opinion, advancing our understanding of the relationship between physical setting and digital engagement.

By assessing sentiments of urban perception and opinion in 2016 and 2022 in the Beijing Second Ring region, this study investigated affective reactions, affective reaction trends, mismatch conditions, and reasons behind the inconsistencies. First, our research built a dataset comprising street view images and social media text posts for human perception and opinion. Next, we constructed the sentiment reaction index, involving image classification (He et al., 2018) based on a questionnaire-based scoring system to measure perception, and natural language processing (NLP) to calculate sentiments in opinions. This research generated two affective reaction maps in both realms during 2016 and 2022, using regression analysis, image segmentation utilizing Vision Transformers for Dense Prediction (DPT) (Ranftl et al., 2021) on ADE20K (Zhou et al., 2017), and word frequency analysis (Neutrino, 2020).

Our results show that the perception affective reaction has more evenly distributed positivity, but increased negative reactions in certain areas, especially around major landmarks. The opinion affective reaction has sharper extremes, while it is particularly positive around key urban areas. For the decrease and increase in sentiment reaction scores, urban elements such as sidewalks, poles, and roads have a significant impact on affective reaction. For example, notable mismatch phenomena are found, especially within the Beijing Second Ring and around major landmarks. In the social media context, topics elicit more polarized shifts in affective reactions. We unveil the affective reaction inconsistency between human perception and opinion about the urban contexts and provide potential explanations for affective reactions for future urban research, environmental management, and urban renewal strategies.

## 2. Methodology

The following framework, shown in Figure 1, utilizes machine learning tools to quantify affective reactions of urban perception through street view images and to measure those opinions by analysing the social media text post dataset.

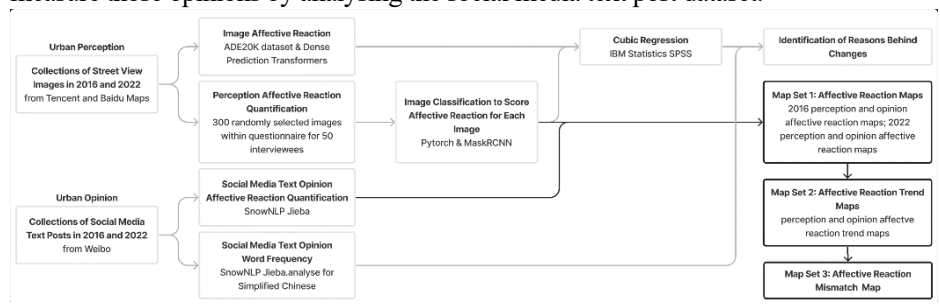

*Figure 1. Affective reaction mapping concept analytical method*

### 2.1. SAMPLE REGION SELECTION

Given the rapid transformations within Chinese society and the intricate urban landscape of Beijing, the second ring area of Beijing's city centre is chosen as the sample region. Therefore, the final investigated region is the square area from (116.343615,39.963175) to (116.460898,39.868876).

### 2.2. DATA COLLECTION AND PROCESSING

To build street view image datasets, we comprised 32,253 geographic-information-tagged images in 2016 captured from Tencent Street View at 900x640 pixels. The street view dataset from 2022 comprised 108,497 images from Baidu Street View, at 4801x320 pixels. Both social media datasets from 2016 and 2022 included 984,024 text posts.

### 2.3. AFFECTIVE REACTION QUANTIFICATION METHOD

To quantify the affective reaction scores of each data, we deployed a questionnaire to request participants to rate the positive or negative feelings they received about the 300 randomly selected images from each of the databases. The collected responses formed the basis of a scoring system that categorized images into 10 score ranges. The categorized images were used to train the model to classify the rest of the images, using MaskRCNN (He et al., 2018) and PyTorch. To reason for the changes, we employed DPT (Ranftl et al., 2021) on MIT's ADE20K (Zhou et al., 2017) for semantic segmentation, to analyse the relationship between urban elements and sentiment reaction scores.

For the social media opinion dataset, the evaluation method uses SnowNLP "Jieba" (Neutrino, 2020), which is architectured for processing simplified Chinese text, to calculate the sentiment score of the textual contents. Furthermore, the most frequently recurring words within these posts were identified, contributing to the understanding of prevailing themes in social media discourse.

2.4. GENERATION OF AFFECTIVE REACTION MAPS

To visualize four affective reaction maps for both urban perception and opinion, a comparative analysis identified and created maps of affective reaction trends (affective reaction in 2022 subtracting that in 2016), by using colour processing in the Grasshopper plugin of Rhino 7. Two affective reaction trend maps are overlaid to generate the affective reaction mismatch map from 2016 to 2022.

## 3. Results and Discussion

Increased neutralization in perception affective reactions might indicate a more balanced urban experience for the physical setting of the city, while heightened extremes in opinion affective reactions from 2016 to 2022 reflect the increasing polarization and public engagement with urban issues, particularly in key infrastructure areas. The following subsections 3.1, 3.2, and 3.3 show affective reaction maps, and subsection 3.4 illustrates the reasons behind the changes.

3.1. AFFECTIVE REACTION MAPPING

From 2016 to 2022, there was a shift towards more evenly distributed positive sentiment, albeit with an increase in negative reactions around major sightseeing places in perception affective reactions.

The 2016 perception affective reaction map, as shown in Figure 2 (a), indicates that people generally felt positive within the second ring while feeling neutral outside this area. The most negative perceptions (scores 0-3) are found around major sightseeing places of the Temple of Heaven, the South area of the Forbidden City, Beiheyan Street, Beichizi Street, and Jingshan Park. The scores of affective perceptions on the Dongsi North Road range from 1 to 8, while the score for the National Art Museum of China on this road is as high as 8.25.

The 2016 opinion affective reaction map, as shown in Figure 2 (b), shows that people felt neutral throughout the investigated area. The most positive affective reactions (scores 9-10) on this map are mainly found around the areas adjacent to the Forbidden City.

The 2022 perception affective reaction map, as shown in Figure 2 (c), indicates that affective reactions were more positive and evenly distributed. The most negative affective reactions (scores below 4) are found on North Long Street along Jingshan Park, Taoranting Park, and the Temple of Heaven (however, the immediately adjacent areas of the Temple of Heaven are as high as 8 to 9).

The 2022 opinion affective reaction map, as shown in Figure 2 (d), indicates a more extreme change in affective reaction: the percentage of the affective reaction score from 0 to 2 increased from 8% to 11.5%, and the percentage of the affective reaction score from 8 to 10 increased from 43% to 57.7%. The most positive affective reactions (scores 9-10) are found around the major urban areas of the Forbidden City, Workers' Stadium, Dongdan, Xidan, and the Temple of Earth.

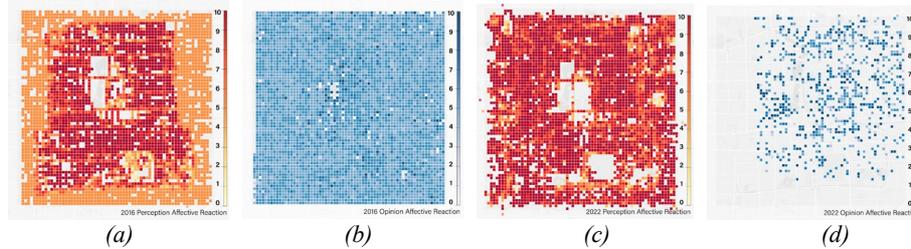

*Figure 2. (a) 2016 perception affective reaction map (the darker the colour, the more positive the affective reaction is); (b) 2016 opinion affective reaction map; (c) 2022 perception affective reaction map; (d) 2016 opinion affective reaction map*

## 3.2. AFFECTIVE REACTION TREND

The affective reaction trend maps show a general decrease in the affective reaction of perception and a more extreme change in the opinions, indicating potentially decreasing affective reaction attachment toward the physical setting and increasing affective reaction expressed on social media.

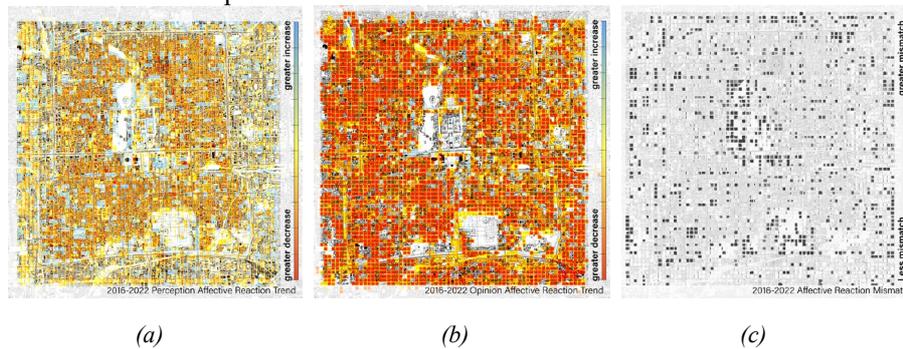

*Figure 3. (a) Perception affective reaction trend map from 2016 to 2022; (b) opinion affective reaction trend map from 2016 to 2022 (red parts indicate a greater decrease in affective reaction and blue parts show a greater increase); (c) affective reaction mismatch map (the blacker parts show a greater mismatch)*

The perception affective reaction trend map indicates a general decrease within the investigated area, while the main increased areas are found around the immediately adjacent regions of main sightseeing areas: the Forbidden City, Tiananmen Square, Jingshan Park, Houhai, Shichahai, the Drum Tower, Dongdan, the Temple of Heaven, and Niu Street. The observed positive shifts in perception around public infrastructures prompt the preference for outdoor areas.

The opinion affective reaction trend map shows a dramatic decrease in affective reaction, while there are a few increased areas: the east area within and adjacent to the Forbidden City, the east area adjacent to Tiananmen Square, Qianmen, the east and west entrance area of the Temple of Heaven. This indicates that the prevailing social media platforms have played a more pivotal role in increasing public engagement online to express opinions, especially under the pandemic impact (Li et al., 2022).

## 3.3. AFFECTIVE REACTION MISMATCH

Significant inconsistencies between urban perceptions and opinions, especially in central areas and around major landmarks suggest a gap between physical urban experiences and their discussions on social media platforms. As shown in Figure 3(c), the most mismatched condition was found at the north part of the central axis in the Forbidden City, areas adjacent to Nanhai, and along Qianmen East Road to the south of Tiananmen Square. The central area experienced clustered mismatch conditions, with notable areas of Shichahai, Beihai, Nanhai, Jingshan Park, the Forbidden City, Tiananmen Square, and the Qianmen area. Temple of Heaven, Longtan Park, and Taoranting Park also experienced more severe affective reaction mismatch conditions.

## 3.4. REASON OF CHANGE

The dataset is districted based on land uses, and then the linear regression analysis by using IBM SPSS Statistics, linking urban elements to street-view perception affective reaction scores yielded illuminating reasons behind the sentiment reaction scores for each zone (including Residential and public Infrastructure, Industry, Storage, External Transportation, Road and Plaza, Municipality, Green, Special, Water and Others, Road). Urban factors include sky, building, green, road, sidewalks, pedestrians (people), transportation, waterbody, seating, fence, sign and symbols, sign lighting, pole, bicyclist, pot, animal, and trash. The subsections selected regression results that have r square values over 0.3 with significances below 0.01, as shown in Figure 4.

**Model Summary and Parameter Estimates**
Dependent Variable: affective reaction

|  | Model Summary | | | | | Parameter Estimates | | |
| --- | --- | --- | --- | --- | --- | --- | --- | --- |
|  | R Square | F | df1 | df2 | Sig. | Constant | b1 | b2 | b3 |
| 2016 Special Zone: Building | .307 | 5.148 | 3 | 59 | .003 | 5.846 | -7.551 | 38.270 | -33.469 |
| 2016 Special Zone: Road | .315 | 5.390 | 3 | 59 | .002 | 8.612 | -31.589 | 109.182 | -132.220 |
| 2016 Special Zone: Pedestrian | .354 | 6.682 | 3 | 59 | <.001 | 5.842 | -99.412 | 10810.378 | -149116.55 |
| 2016 ExtTrans Zone: Pedestrian | .395 | 2.591 | 3 | 32 | .007 | 5.048 | 1.763 | 1.502 | 47.661 |
| 2022 Special Zone: Pole | .305 | 8.644 | 3 | 59 | <.001 | 8.010 | -195.604 | 483993.55 | -1202694 |
| 2022 Special Zone: Pot | .308 | 13.349 | 2 | 60 | <.001 | 8.082 | .000 | 155308.72 | -18417388 |
| 2022 Other Zone: Sidewalk | .337 | 5.918 | 3 | 35 | .002 | 8.078 | -57.045 | 725.762 | -2034.725 |
| 2022 Storage Zone: Sidewalk | .307 | 2.512 | 3 | 17 | .009 | 8.709 | -120.126 | 1407.487 | -4113.280 |
| 2022 Storage Zone: Pedestrian | .307 | 2.511 | 3 | 17 | .009 | 7.648 | 750.355 | -504106.9 | 31568801 |
| 2022 Storage Zone: Pole | .359 | 3.180 | 3 | 17 | .001 | 7.769 | -1344.412 | 184503.57 | -57821118.0 |

*Figure 4. Model summary and parameters for image segmentation and affective reaction.*

### 3.4.1. 2016 Affective Reaction Reasoning

As shown in Figure 5(a), people felt more positive towards the rising proportion of buildings until its proportion reached 60% of an image; in Figure 5(b), people felt more negative towards the more road presence; in Figure 5(c), people felt more positive towards the rising proportion of pedestrian people.

INCONSISTENT AFFECTIVE REACTIONS:
Sentiment of Perception and Opinion in Urban Environments

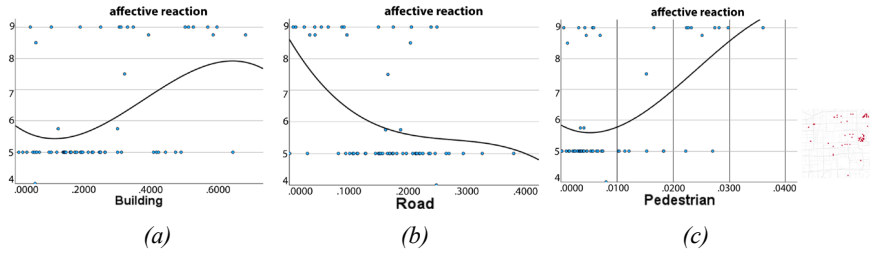

(a)          (b)          (c)

*Figure 5. (a) Cubic regression analysis between sentiment reactions and building segment proportion in the street view images in 2016; (b) cubic regression between sentiment and road segment proportion; (c) cubic regression between sentiment and pedestrian segment proportion in Special zoning category*

For external transportation zoning, as shown in Figure 6, people felt more positive toward the rising proportion of roads in a street view image.

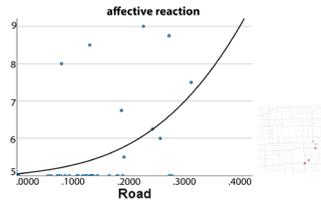

*Figure 6. Cubic regression analysis between sentiment reactions and road segment proportion in the street view images in 2016 in the External Transportation zoning category.*

### 3.4.2. 2022 Affective Reaction Reasoning

As shown in Figure 7 (a) and Figure 7 (b), people felt negatively about the poles and pots in images in the Special Zoning category in 2022 Perception Sentiment Reactions. In the Other zoning shown in Figure 7 (c), people felt positive about the rising portion of the sidewalk until it reached 20% of the image.

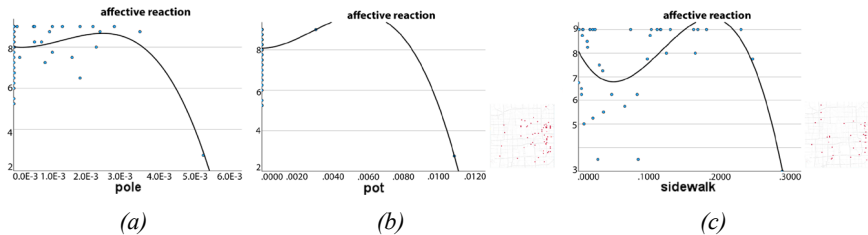

(a)          (b)          (c)

*Figure 7. (a) Cubic regression analysis between sentiment reactions and pole segment proportion in the street view images in 2022; (b) cubic regression between sentiment reactions and pot proportion in 2022 in Special zoning category; (c)cubic regression analysis between sentiment reactions and sidewalk segment proportion in the street view images in 2022 in Other zoning category*

As shown in Figure 8, in the Storage zoning area, people felt increasingly negative towards the rising proportion of the sidewalk after it reached 20% of the image, felt increasingly positive towards the rising proportion of pedestrians, and felt increasingly negative towards the rising proportion of the visible pole.

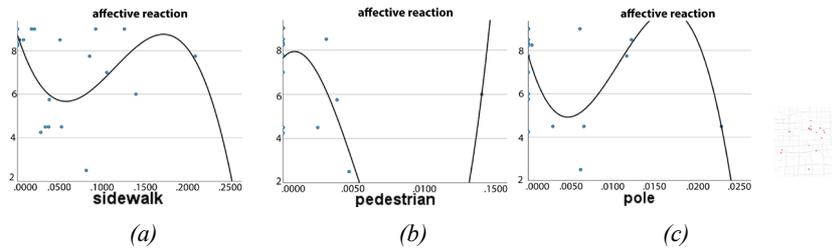

*(a)* *(b)* *(c)*

*Figure 8. (a) Cubic regression analysis between sentiment reactions and sidewalk segment proportion in the street view images in 2022; (b) cubic regression analysis between sentiment and pedestrian proportion; (c) cubic regression between sentiment reactions and pole proportion in Storage zoning category*

In summary, the positive reaction to buildings up to 60% suggests a preference for moderately dense urban environments, while the positive sentiment toward pedestrian presence highlights the preference for walkable, lively streetscapes (Gerike et al., 2021; Kim et al., 2017). Excessive road coverage is generally perceived negatively, possibly due to associated congestion and environmental concerns (Ma et al., 2021). Negative reactions to signs, symbols, and other elements like poles and pots suggest public preference for more accessible, less cluttered, visually more appealing urban spaces.

### 3.4.3. Social Media Affective Reaction Reasoning

NLP analysis extracted frequent words in the posts. In 2016, terms such as "Beijing," "share," "photo," and "no" dominated, while in 2022, "Beijing" and "supertopic" emerged as prominent keywords. The change from general sharing ("share" and "photo") to "supertopic" likely reflects the evolution of social media platforms towards more organized and thematic content, possibly influenced by algorithmic changes and user behavior trends visible on the social media platforms (Bonini & Gandini, 2019), leading to more polarized affective reaction changes (X. Zhang et al., 2021), indicating how algorithm-curated content can shape certain views and affective reactions. Regulatory changes on social media platforms could influence the types of discussions and the language used, possibly contributing to the shift toward more controlled discourse (Tsoy et al., 2021).

## 4. Conclusion

Our research focuses on the relationship between the urban environment and affective reactions embedded within human perception and opinion; it proposes a new approach to quantifying, on a large scale, affective reactions, mismatched conditions, and reasons behind these phenomena based on analysing street view images and text posts. This methodology to measure mismatched reactions can help future research based on street view and social media data, especially large-scale urban affective reaction analysis, to further adjust and explain the bias in research results caused by street view and social media data. Our results suggest the need for balanced urban density and pedestrian-friendly designs for perception affective reactions within urban environments. The polarization shift in opinion affective reaction especially around large infrastructure and major landmarks, indicates more government engagements are necessary for effective communication with the public on digital platforms.

These analytical methods are at an early stage with several limitations and possibilities for improvement. This study's reliance on visual data from street views and textual data from social media may not fully capture the multi-dimensional expression of urban reactions. Specific to Beijing's urban context and covering a specific timeframe from 2016 to 2022, our results may not apply to different cultural and urban settings beyond this period.

Future improvement could be made by including more qualitative data or applying affective reaction detection directly over facial expressions to improve the accuracy of affective reaction quantification. Further development could be carried to additional cities with broader time spans, to explore more accurate relationships between affective reaction, urban context, and time.

**References**

Alaei, A., Becken, S., & Stantic, B. (2017). *Sentiment analysis in tourism: Capitalising on Big Data*. https://doi.org/10.1177/0047287517747753

Becken, S., Stantic, B., Chen, J., Alaei, A. R., & Connolly, R. M. (2017). Monitoring the environment and human sentiment on the Great Barrier Reef: Assessing the potential of collective sensing. *Journal of Environmental Management*, *203*, 87–97. https://doi.org/10.1016/j.jenvman.2017.07.007

Beyes, T., Holt, R., & Pias, C. (Eds.). (2019). *The Oxford Handbook of Media, Technology, and Organization Studies* (1st ed.). Oxford University Press. https://doi.org/10.1093/oxfordhb/9780198809913.001.0001

Biljecki, F., & Ito, K. (2021). Street view imagery in urban analytics and GIS: A review. *Landscape and Urban Planning*, *215*, 104217. https://doi.org/10.1016/j.landurbplan.2021.104217

Bonini, T., & Gandini, A. (2019). "First Week Is Editorial, Second Week Is Algorithmic": Platform Gatekeepers and the Platformization of Music Curation. *Social Media + Society*, *5*(4), 2056305119880000. https://doi.org/10.1177/2056305119880006

Dai, Z., & Hua, C. (2019). The Improvement of Street Space Quality Measurement Method Based on Streetscape. *Planners*, 57–63.

Gerike, R., Koszowski, C., Schröter, B., Buehler, R., Schepers, P., Weber, J., Wittwer, R., & Jones, P. (2021). Built Environment Determinants of Pedestrian Activities and Their Consideration in Urban Street Design. *Sustainability*, *13*(16), 9362. https://doi.org/10.3390/su13169362

He, K., Gkioxari, G., Dollár, P., & Girshick, R. (2018). *Mask R-CNN* (arXiv:1703.06870). arXiv. http://arxiv.org/abs/1703.06870

Kim, T., Sohn, D.-W., & Choo, S. (2017). An analysis of the relationship between pedestrian traffic volumes and built environment around metro stations in Seoul. *KSCE Journal of Civil Engineering*, *21*(4), 1443–1452. https://doi.org/10.1007/s12205-016-0915-5

Li, K., Zhou, C., Luo, X. (Robert), Benitez, J., & Liao, Q. (2022). Impact of information timeliness and richness on public engagement on social media during COVID-19 pandemic: An empirical investigation based on NLP and machine learning. Decision Support Systems, 162, 113752. https://doi.org/10.1016/j.dss.2022.113752

Ma, Y., Yang, Y., & Jiao, H. (2021). Exploring the Impact of Urban Built Environment on Public Emotions Based on Social Media Data: A Case Study of Wuhan. *Land*, *10*(9), 986. https://doi.org/10.3390/land10090986

McLuhan, M. (1964). *Understanding Media*.

Naik, N., Philipoom, J., Raskar, R., & Hidalgo, C. (2014). Streetscore—Predicting the Perceived Safety of One Million Streetscapes. *2014 IEEE Conference on Computer Vision and Pattern Recognition Workshops*, 793–799. https://doi.org/10.1109/CVPRW.2014.121

Neutrino. (2020). *"Jieba" (Chinese for "to stutter") Chinese text segmentation: Built to be the best Python Chinese word segmentation module.* https://github.com/fxsjy/jieba

Oungrinis, K. (2006). *Transformations: Paradigms for designing transformable spaces*. Cambridge, MA : Dept. of Architecture, Harvard Design School [sic].

Pei, Y., Kan, C., & Dang, A. (2019). Street Greenspace Justice Assessment Study of Dongcheng District in Beijing Based on Street View Data. *Chinese Landscape Architecture*. https://doi.org/10.19775/j.cla.2020.11.0051

Ranftl, R., Bochkovskiy, A., & Koltun, V. (2021). *Vision Transformers for Dense Prediction* (arXiv:2103.13413). arXiv. http://arxiv.org/abs/2103.13413

Shook, E., & Turner, V. K. (2016). The socio-environmental data explorer (SEDE): A social media–enhanced decision support system to explore risk perception to hazard events. *Cartography and Geographic Information Science*, *43*(5), 427–441. https://doi.org/10.1080/15230406.2015.1131627

Tao, Y., Zhang, F., Shi, C., & Chen, Y. (2019). Social Media Data-Based Sentiment Analysis of Tourists' Air Quality Perceptions. *Sustainability*, *11*(18), 5070. https://doi.org/10.3390/su11185070

Tsoy, D., Tirasawasdichai, T., & Ivanovich Kurpayanidi, K. (2021). Role of Social Media in Shaping Public Risk Perception during COVID-19 Pandemic: A Theoretical Review. *THE INTERNATIONAL JOURNAL OF MANAGEMENT SCIENCE AND BUSINESS ADMINISTRATION*, *7*(2), 35–41. https://doi.org/10.18775/ijmsba.1849-5664-5419.2014.72.1005

van der Meer, T. G. L. A., & Hameleers, M. (2022). I Knew It, the World is Falling Apart! Combatting a Confirmatory Negativity Bias in Audiences' News Selection Through News Media Literacy Interventions. *Digital Journalism*, *10*(3), 473–492. https://doi.org/10.1080/21670811.2021.2019074

Winner, L. (2023). *Do Artifacts Have Politics?*

Yazıcıoğlu, D. (2017). *THE SMARTPHONE AFFECT: The Emotional Impact of Smartphone Usage in Public Spaces and it's Affects on the Subjective Experience of Public Space's Sociality*. https://doi.org/10.13140/RG.2.2.32888.90882

Ye, Y., Yin, R., Hu, Y., & Fang, J. (2021). Measuring the Impacts of Fine-Scale Urban Forms on Street Temperatures and Design Responses. *Landscape Architecture*. https://doi.org/10.14085/j.fjyl.2021.08.0058.08

Zhang, F., Zhou, B., Liu, L., Liu, Y., Fung, H. H., Lin, H., & Ratti, C. (2018). Measuring human perceptions of a large-scale urban region using machine learning. *Landscape and Urban Planning*, *180*, 148–160. https://doi.org/10.1016/j.landurbplan.2018.08.020

Zhang, X., Chen, A., & Zhang, W. (2021). Before and after the Chinese gene-edited human babies: Multiple discourses of gene editing on social media. *Public Understanding of Science*, *30*(5), 570–587. https://doi.org/10.1177/0963662520987754

Zhou, B., Zhao, H., Puig, X., Fidler, S., Barriuso, A., & Torralba, A. (2017). Scene Parsing through ADE20K Dataset. *2017 IEEE Conference on Computer Vision and Pattern Recognition (CVPR)*, 5122–5130. https://doi.org/10.1109/CVPR.2017.544